\documentclass[conference]{IEEEtran}
\IEEEoverridecommandlockouts
\usepackage{cite}
\usepackage{url}
\usepackage{amsmath,amssymb,amsfonts}
\usepackage{algorithmic}
\usepackage{graphicx}
\usepackage{textcomp}
\usepackage{xcolor}
\usepackage{verbatim}
\def\BibTeX{{\rm B\kern-.05em{\sc i\kern-.025em b}\kern-.08em
    T\kern-.1667em\lower.7ex\hbox{E}\kern-.125emX}}
\begin{document}

\title{Tailoring Generative Adversarial Networks for Smooth Airfoil Design \\
}

\makeatletter
\newcommand{\linebreakand}{%
  \end{@IEEEauthorhalign}
  \hfill\mbox{}\par
  \mbox{}\hfill\begin{@IEEEauthorhalign}
}
\makeatother


\author{\IEEEauthorblockN{Joyjit Chattoraj}
\IEEEauthorblockA{\footnotesize \textit{Institute of High Performance Computing} \\
\textit{\footnotesize Agency for Science, Technology and Research}\\
Singapore 138632, Republic of Singapore\\
joyjitc@ihpc.a-star.edu.sg}
\and
\IEEEauthorblockN{Jian Cheng Wong}
\IEEEauthorblockA{\footnotesize \textit{Institute of High Performance Computing} \\
\textit{\footnotesize Agency for Science, Technology and Research}\\
Singapore 138632, Republic of Singapore\\
wongj@ihpc.a-star.edu.sg}
\and
\IEEEauthorblockN{Zhang Zexuan}
\IEEEauthorblockA{\footnotesize \textit{School of Continuing and Lifelong Education} \\
\textit{\footnotesize National University of Singapore}\\
Singapore 119260, Republic of Singapore\\
E0954403@u.nus.edu}
\linebreakand
\IEEEauthorblockN{Manna Dai}
\IEEEauthorblockA{\footnotesize \textit{Institute of High Performance Computing} \\
\textit{\footnotesize Agency for Science, Technology and Research}\\
Singapore 138632, Republic of Singapore\\
manna\_dai@ihpc.a-star.edu.sg}
\and
\IEEEauthorblockN{Xia Yingzhi}
\IEEEauthorblockA{\footnotesize \textit{Institute of High Performance Computing} \\
\textit{\footnotesize Agency for Science, Technology and Research}\\
Singapore 138632, Republic of Singapore\\
xia\_yingzhi@ihpc.a-star.edu.sg}
\and
\IEEEauthorblockN{Li Jichao}
\IEEEauthorblockA{\footnotesize \textit{Institute of High Performance Computing} \\
\textit{\footnotesize Agency for Science, Technology and Research}\\
Singapore 138632, Republic of Singapore\\
li\_jichao@ihpc.a-star.edu.sg}
\linebreakand
\IEEEauthorblockN{Xu Xinxing}
\IEEEauthorblockA{\footnotesize \textit{Institute of High Performance Computing} \\
\textit{\footnotesize Agency for Science, Technology and Research}\\
Singapore 138632, Republic of Singapore\\
xuxinx@ihpc.a-star.edu.sg}
\and
\IEEEauthorblockN{Ooi Chin Chun}
\IEEEauthorblockA{\footnotesize \textit{Institute of High Performance Computing} \\
\textit{\footnotesize Agency for Science, Technology and Research}\\
Singapore 138632, Republic of Singapore \\
ooicc@cfar.a-star.edu.sg}
\and
\IEEEauthorblockN{Yang Feng}
\IEEEauthorblockA{\footnotesize \textit{Institute of High Performance Computing} \\
\textit{\footnotesize Agency for Science, Technology and Research}\\
Singapore 138632, Republic of Singapore \\
yangf@ihpc.a-star.edu.sg}
\linebreakand
\IEEEauthorblockN{Dao My Ha}
\IEEEauthorblockA{\footnotesize \textit{Institute of High Performance Computing} \\
\textit{\footnotesize Agency for Science, Technology and Research}\\
Singapore 138632, Republic of Singapore \\
daomh@ihpc.a-star.edu.sg}
\and
\IEEEauthorblockN{Liu Yong}
\IEEEauthorblockA{\footnotesize \textit{Institute of High Performance Computing} \\
\textit{\footnotesize Agency for Science, Technology and Research}\\
Singapore 138632, Republic of Singapore \\
liuyong@ihpc.a-star.edu.sg}
}

\maketitle
\thispagestyle{plain}
\pagestyle{plain}

\begin{abstract} 
In the realm of aerospace design, achieving smooth curves is paramount, particularly when crafting objects such as airfoils. Generative Adversarial Network (GAN), a widely employed generative AI technique, has proven instrumental in synthesizing airfoil designs. However, a common limitation of GAN is the inherent lack of smoothness in the generated airfoil surfaces. To address this issue, we present a GAN model featuring a customized loss function built to produce seamlessly contoured airfoil designs. Additionally, our model demonstrates a substantial increase in design diversity compared to a conventional GAN augmented with a post-processing smoothing filter.
\end{abstract}

\begin{IEEEkeywords}
GAN, generative AI, custom loss, airfoil design
\end{IEEEkeywords}

\section{Introduction}\label{Sec:Intro}
Generative AI models, such as Generative Adversarial Network (GAN) and Variational Autoencoder (VAE), have demonstrated promising outcomes in engineering design. Applications span a range of fields, encompassing structural optimization, materials design, and shape synthesis~\cite{regenwetter2022deep}.
In recent years, particularly in the domain of airfoil design~\cite{Li2022}, numerous variations of GAN models have been investigated to discover novel shapes capable of achieving optimal aerodynamic performance across a diverse set of operating conditions~\cite{li2020efficient, yilmaz2020conditional, heyrani2021pcdgan, Li2021a, du2022airfoil}.

However, it is frequently reported in the literature that GAN has limitations generating smooth airfoil curves~\cite{achour2020development,  tan2022airfoil, santos2023using, wada2023physics}. 
A commonly employed strategy to address this issue involves integrating a smoothing filter as a post-processing technique~\cite{achour2020development}.

Chen et al~\cite{chen2018b,chen2019aerodynamic,chen2020airfoil} proposed a novel model, B{\'e}zierGAN, to avoid post-processing of airfoil curves, instead, they employed a smoothing method, in this case, B{\'e}zier curve, as an integral part of the GAN generator. During GAN training, the model learns the optimized values for B{\'e}zier parameters.      
In addition, a set of regularization terms has been imposed that enhances the complexity of the model, which could potentially limit the model's generalizability across various applications.

Yonekura et al~\cite{yonekura2022inverse} used Conditional Wasserstein GAN (CWGAN-GP) 
and found that the model can generate smooth airfoil curves. Notably, WGAN was originally proposed to tackle the vanishing gradient problem in GAN models~\cite{arjovsky2017wasserstein}. 
Hence, whether the primary reason for non-smoothness lies in a vanishing gradient remains an open question.
Tan et al~\cite{tan2022airfoil} with a similar CWGAN-GP model could not generate smooth airfoil curves consistently. 
One noticeable difference between these two models is that Tan et al~\cite{tan2022airfoil} considered more than one parameter for airfoil design including a shape-determined parameter (area).          

Wang et al~\cite{wang2023airfoil} achieved smooth airfoil shapes by integrating a VAE and a GAN, referred to as VAEGAN, originally proposed in~\cite{larsen2016autoencoding}.
Here, VAE acts as a GAN generator, and it can generate smooth curves without imposing any explicit methodology for curve smoothness. 

\begin{figure}[t]
\centerline{\includegraphics[width=1\linewidth]{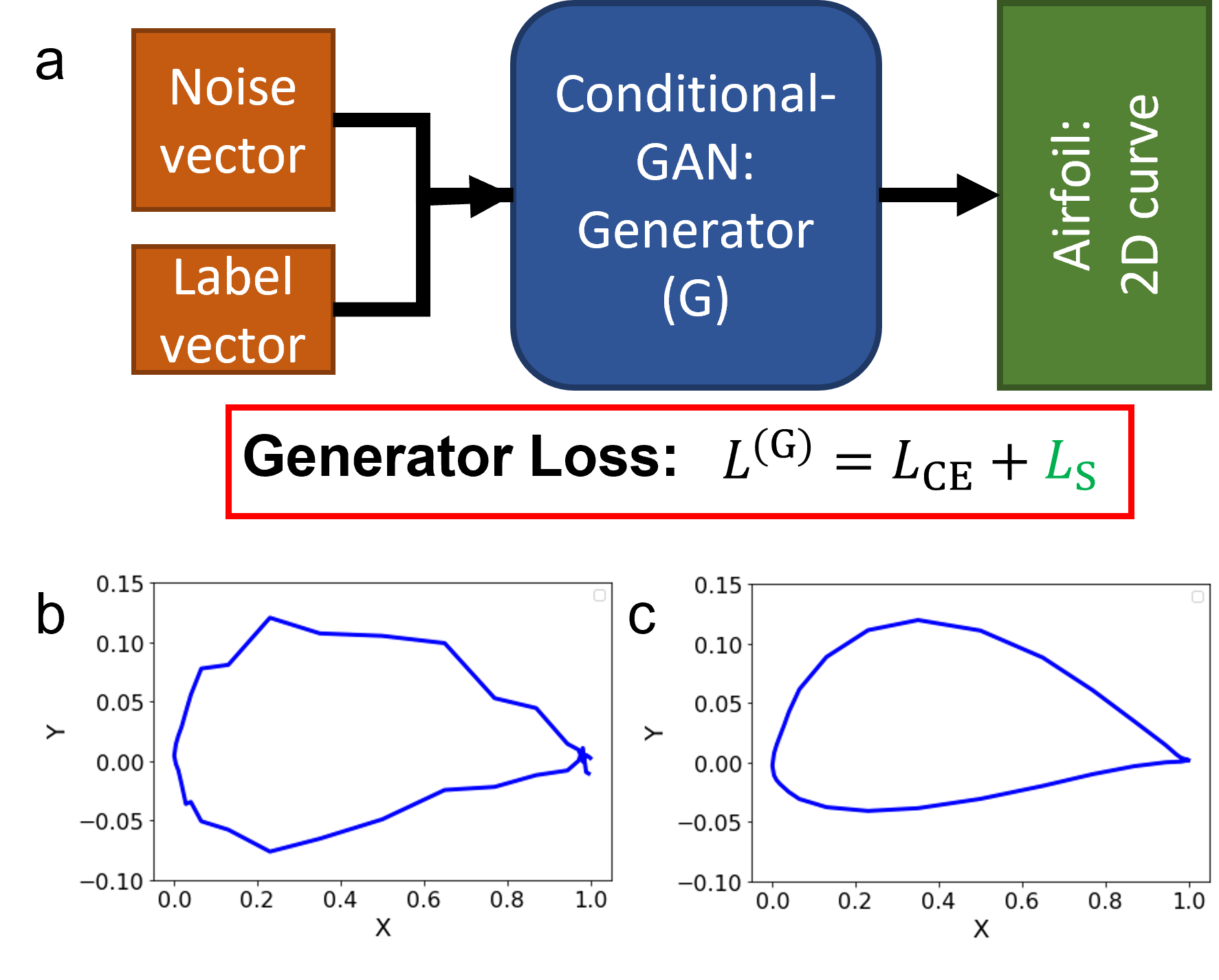}}
\caption{(a) The generator (G) of a trained conditional-GAN takes a noise vector and a label vector as inputs and the generator outputs a 2D airfoil curve. The generator is trained with a loss function $L^{(\text{G})}$ which includes a binary cross entropy loss ($L_{\text{CE}}$) and a mean square error loss $L_{\text{S}}$ to impose curve smoothing. (b) The generator outputs non-smooth curves if $L^{(\text{G})}$ only includes $L_{\text{CE}}$, i.e., without $L_{\text{S}}$. (c) The generator outputs desired smooth curves when $L_{\text{S}}$ is included in the generator loss in addition to $L_{\text{CE}}$.}
\label{Fig:G-Loss}
\end{figure}

In our proposed methodology, we attempt to tackle the non-smoothness problem of GAN explicitly. We propose a custom loss function for the generator model of conditional GAN (Fig.~\ref{Fig:G-Loss}a). This loss function calculates a mean square error if a generated curve deviates from its moving average position. The custom loss function improves the ability of the GAN model to transition from generating non-smooth curves to creating smooth airfoil curves, as illustrated in Fig.~\ref{Fig:G-Loss}b,c. Moreover, we find that the custom loss facilitates the GAN model to generate more diverse airfoil shapes than the original GAN without the custom loss function and incorporated with a smoothing filter as a post-processing technique. As the custom loss requires computing only the moving average, this approach is generalizable and can be easily adopted in other engineering designs to impose smoothness.  

The remaining sections of this article are organized as follows: (i) Dataset: Here, we provide an overview of the airfoil data utilized for training the GAN models.
(ii) Methodology: This section delves into the specifics of the GAN architecture, custom loss, and the training procedure. (iii) Results: We analyze the effectiveness of the custom loss function in this segment. (iv) Conclusions: Finally, we summarize the main findings and discuss potential avenues for future research.

\section{Dataset}
Our dataset consists of 1399 airfoils which are obtained from the UIUC Airfoil Data Site~\cite{selig1996uiuc} using an open-source pre-processing code~\cite{Li2019,li2019data}. These airfoils are essentially 2D curves with a set of X and Y coordinates, where X coordinates strictly range from 0 to 1. The Y coordinates are flexible and typically vary between $-$0.3 to 0.4 (Fig.~\ref{Fig:p38}).
The maximum thickness ($\tau$), i.e., measured perpendicular to the camber line of an airfoil, varies between 0.04 and 0.38 with a median value of 0.12. 

Subsequently, these airfoils underwent assessment under three distinct operating conditions: Reynolds number ($Re$), Mach number ($M$), and angle of attack ($\alpha$). The airfoil's aerodynamic performance was evaluated by calculating the lift coefficient ($c_l$) and drag coefficient ($c_d$) utilizing XFOIL~\cite{drela1989xfoil}. The numerical simulation generated a comprehensive dataset comprising 654,963 samples, averaging 468 samples per airfoil, encompassing different operating conditions, i.e., $Re \ (2000000 - 12000000)$, $M \ (0 - 0.3)$, and $\alpha \ (0 - 33)$, with their calculated $c_l$ and $c_d$.

\begin{figure}[t]
\centerline{\includegraphics[width=0.95\linewidth]{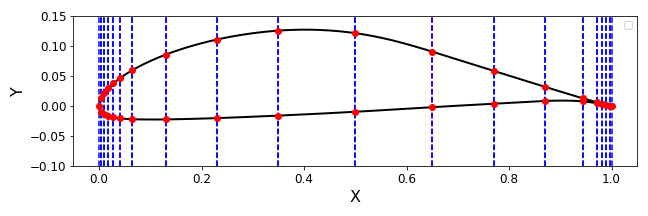}}
\caption{An original airfoil (solid line) after reducing the number of points to 38 (symbols).}
\label{Fig:p38}
\end{figure}
\begin{table}[htbp]
\caption{8 Classes and respective number of airfoils for training }
\begin{center}
\begin{tabular}{|c|c|c|c|c|}
\hline
Class & $c_l/c_d$ & $\alpha$ & $\tau$ & Airfoils \\
\hline
000 & low  & low  & low  &  740\\
001 & low  & low  & high &  627\\
010 & low  & high & low  &  742\\
011 & low  & high & high &  649\\
100 & high & low  & low  &  739\\
101 & high & low  & high &  633\\
110 & high & high & low  &  717\\
111 & high & high & high &  566\\
\hline
\end{tabular}
\label{Table:class}
\end{center}
\end{table}
In essence, our GAN design challenge is framed as a classification problem, where the objective is to create airfoil shapes that align with a particular class.
We define 8 classes to create airfoil shapes from 3 parameters: $\tau$ representative of airfoil geometry, $\alpha$ representative of operating conditions, and the ratio $c_l/c_d$ representative of aerodynamic performance (Table~\ref{Table:class}). 
A class is characterized by combinations of low and high values of $\tau$, $\alpha$, and $c_l/c_d$. The thresholds for distinguishing between low and high values are established based on the median values of these parameters, which are 0.12, 10, and 100, respectively. 
We proceed to create a training dataset by determining the class to which each airfoil belongs. Note that while an airfoil may be associated with multiple classes, its presence in a class is considered only once.
We find that the number of airfoils varies between 566 and 742 in 8 classes, a total of 5413 samples, implying a reasonably balanced dataset for GAN training. 

\section{Methodology}
\begin{figure*}[t]
\centerline{\includegraphics[width=0.9\linewidth]{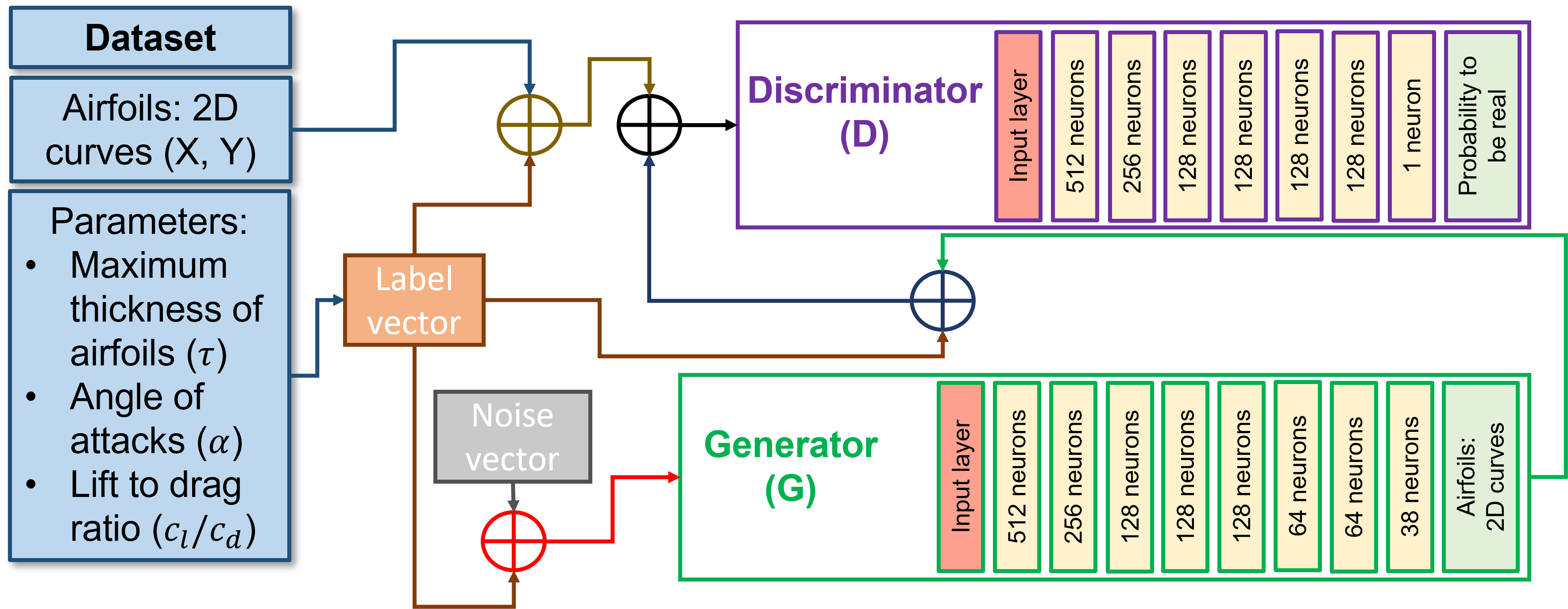}}
\caption{The architecture of conditional-GAN consists of a generator (G) and a discriminator (D). Both G and D are fully connected deep neural networks.}
\label{Fig:GAN}
\end{figure*}
\subsection{Pre-processing} 
We employ a pre-processing methodology to standardize airfoil coordinates for GAN training. This process ensures uniform X coordinates across all airfoils, maintaining consistency in their sequence, while strategically reducing the number of points. Emphasis is placed on critical airfoil segments, such as the leading and trailing edges. Employing a cosine distribution, we set 20 X coordinates; excluding the two endpoints (X=0 and X=1), the remaining 18 X coordinates each have 2 corresponding Y coordinates for the upper and lower surfaces (Fig.~\ref{Fig:p38}). Consequently, a total of 38 Y coordinates distinguish one airfoil from another, and these 38 Y coordinates serve as inputs for GAN training.

\subsection{GAN architecture}
We adopt the conditional GAN technique~\cite{mirza2014conditional} which comprises a generator (G) and a discriminator (D), depicted in Fig.~\ref{Fig:GAN}. G is a fully connected deep neural network with the rectified linear unit (ReLU) as an activated function. 
G receives an input vector formed by concatenating a random noise vector (128 dimensions, Gaussian noise with mean 0 and standard deviation 1) and a label vector (6 dimensions).
Label vector represents the 8 classes (Table~\ref{Table:class}), with an example being 100101 label vector for class-011. G creates 38 Y coordinates belonging to the 2D curve of an airfoil. 

D receives an input vector formed by concatenating the real coordinates (38 dimensions) and generated coordinates (38 dimensions) from G associated with the label vector (6 dimensions). D determines the probability of the generated coordinates to be real. D is also a fully connected deep neural network; the last layer of D uses the sigmoid activation function, and all the other neurons use ReLU. 

\subsection{Custom loss function}
To enforce smoothness on the 2D curve of an airfoil,  we tailor the generator training by customizing the loss function $L^\text{(G)}$ as   
\begin{equation}
    L^\text{(G)} = L_\text{CE} + L_\text{S},
    \label{Eq:Lg}
\end{equation}
where $L_\text{CE}$ is the standard binary cross-entropy loss, whereas $L_\text{S}$ introduces a penalty for any deviation of the generated Y coordinates (with dimensions $n_c=38$) from their respective moving average points.     
We compute the moving average $\bar{Y}_i$ for all Y coordinates with a window of 3 points as 
\begin{equation}
    \bar{Y}_i = \frac{1}{3}\sum_{j=-1}^1 Y_{i+j},
\end{equation}
where $i=1, 2, \ldots, n_c$, and the index $i+j$ iterates through 1 and $n_c$ in a cyclic manner. 
The coordinates are arranged in a loop where $i=1$ corresponds to the endpoint at $X_i=1$, $i=2$ corresponds to the upper surface point at $X_i=0.996$, and $i=n_c$ corresponds to the lower surface point at $X_i=0.996$.   
The deviation $\Delta_i$ between a Y coordinate $Y_i$ and its moving average point is then   
\begin{equation}
    \Delta_i = Y_i - \bar{Y}_i. 
\end{equation}
We define the smoothing loss $L_\text{S}$ as the mean square deviation of $\Delta_i$ as shown below
\begin{equation}
    L_{\text{S}} = \frac{\omega}{n_c}\sum_{i=1}^{n_c}\Delta_i^2,
    \label{Eq:Ls}
\end{equation}
where the coefficient $\omega$ determines the strength of $L_\text{S}$ with respect to $L_\text{CE}$. 
For the discriminator, we use the binary cross-entropy loss function only.

\subsection{GAN training}
The generator (G) and discriminator (D) are trained simultaneously in a competitive manner incorporating the custom loss function~(\ref{Eq:Lg}). In a training loop, G creates 2D airfoil curves, and D evaluates both real and generated airfoils without knowing the source. 
For both G and D, we use ADAM optimizer and the learning rate $10^{-4}$. We set epochs at $30,000$.
During training we observed that beyond the first several thousand epochs, both the G-loss and  D-loss converged to approximately the same value, maintaining a steady-state until the end of the training process.
We experimented with three distinct values for $\omega$ in (\ref{Eq:Ls}): 1, 10, and 100. Our findings indicated that setting $\omega$ to 10 resulted in optimal smoothing for the airfoils. 

\section{Results}
\begin{figure}[t]
\centerline{\includegraphics[width=0.9\linewidth]{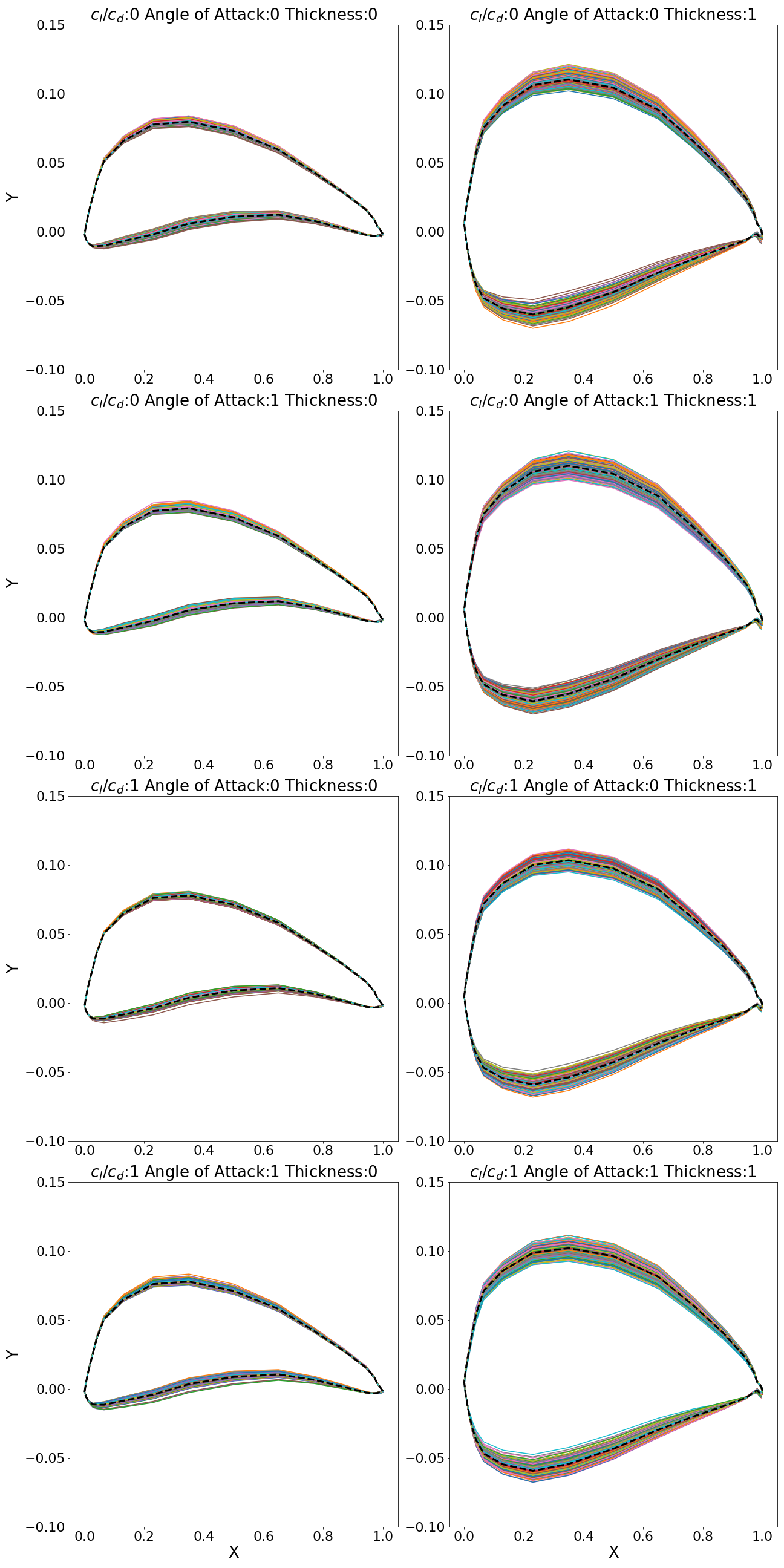}}
\caption{Generated airfoil samples from the GAN trained without the smoothing loss and augmented with a post-processing smoothing filter. Each panel represents a class defined by low/high (0/1) $c_l/c_d$, $\alpha$, and $\tau$. It displays 600 airfoil samples and their mean shape (dashed curve).}
\label{Fig:AirfolisWoSmoothing}
\end{figure}

To evaluate the efficacy of our proposed custom loss function (\ref{Eq:Lg}), we conducted training experiments with two GAN models: one without a smoothing loss and another incorporating the smoothing loss function, to be referred to as smoothGAN, henceforth. We then compared their performances in generating airfoils. A total of 600 airfoil samples were generated for each of the 8 classes, and we assessed these samples using three statistical metrics:
\begin{enumerate}
\item $\text{ACC}^{(\tau)}$: Accuracy that measures the percentage of the total airfoils possessing the correct (low/high) thickness at a given class.
\item $\sigma^{(\tau)}$: Diversity in thickness, calculated as the standard deviation of thickness after being rescaled by their maximum observed value.
\item $S$: Diversity in shape, following~\cite{brown2019quantifying}, it is calculated as the average distance of the mean shape $Y^\text{(m)}$ from all the generated samples
$\left\langle \sqrt{\sum_{i=1}^{n_c} (Y_i - Y_i^\text{(m)})^2} \right\rangle_\text{samples}$.
\end{enumerate}
\begin{figure}[t]
\centerline{\includegraphics[width=0.9\linewidth]{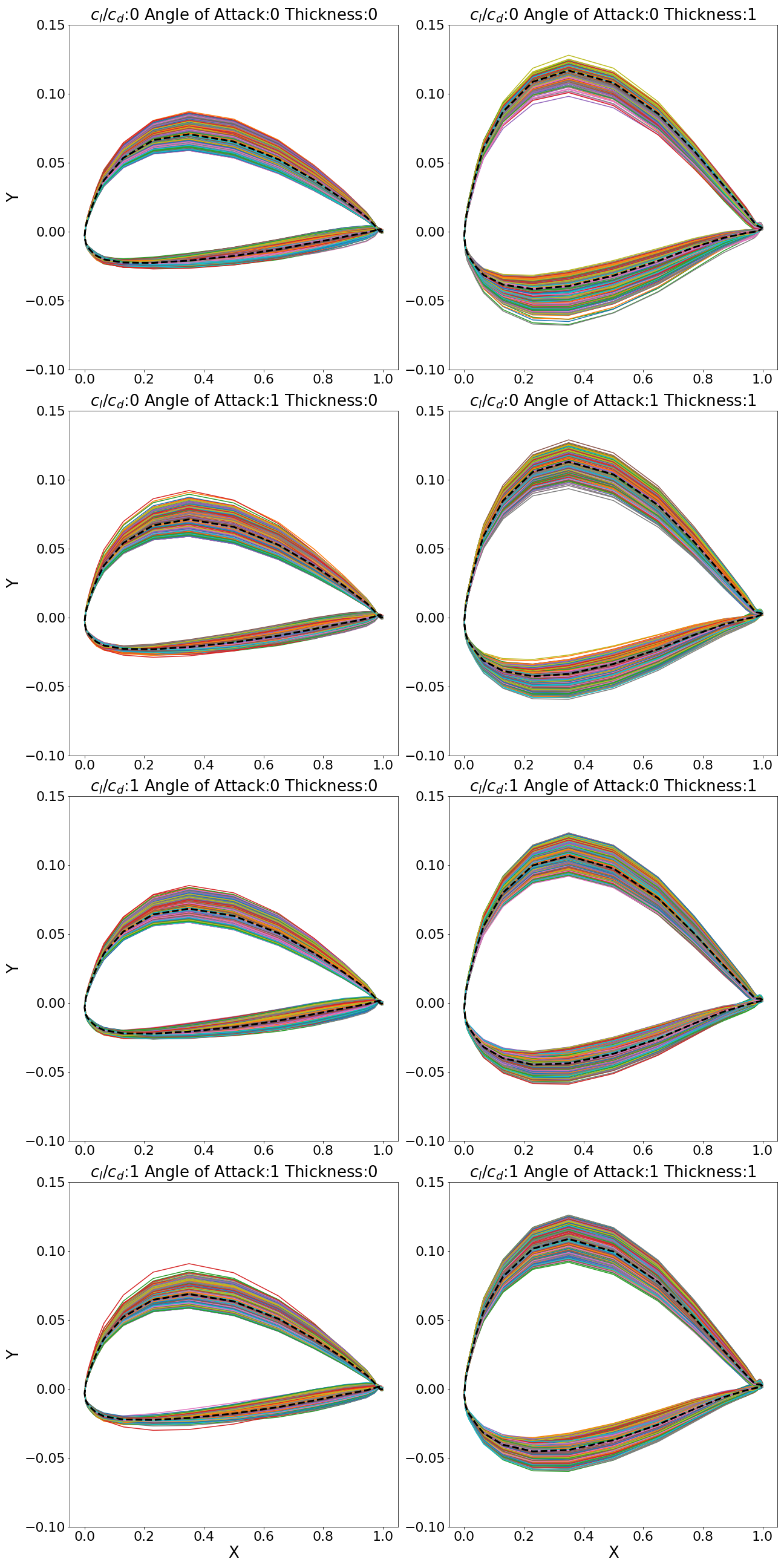}}
\caption{Generated airfoil samples from the GAN model trained with the smoothing loss. Each panel represents a class defined by low/high (0/1) $c_l/c_d$, $\alpha$, and $\tau$. It displays 600 airfoil samples and their mean shape (dashed curve).}
\label{Fig:AirfolisWSmoothing}
\end{figure}

\subsection{GAN without smoothing loss function}
We found that GAN which is trained without the smoothing loss function always generated non-smooth 2D curves, an example can be seen in Fig.~\ref{Fig:G-Loss}b.
A Savitzky‐Golay filter~\cite{press1990savitzky} is further augmented with the GAN model as a post-processing technique that smooths out the GAN-generated samples, which are displayed in Fig.~\ref{Fig:AirfolisWoSmoothing}.
We illustrate the mean shape of airfoils for each class in Fig.~\ref{Fig:Results}a to showcase the shape diversity among classes. Additionally, the distribution of $\tau$ is depicted in Fig.~\ref{Fig:Results}c,e as an alternative measure of geometric diversity.

\subsection{smoothGAN: GAN with smoothing loss function}
The smoothGAN model trained with the custom loss function generates smooth 2D curves, which do not require any post-processing. 600 samples for each of 8 classes, a total of 4800, plotted in Fig.~\ref{Fig:AirfolisWSmoothing}, all exhibits smooth shapes. 
It is evident in Fig.~\ref{Fig:AirfolisWSmoothing} and also in Fig.~\ref{Fig:Results}b,d,f that in comparison to GAN, smoothGAN achieves more diversity in generated shapes. We further quantitatively demonstrate the diversity through two metrics: $\sigma^{(\tau)}$ and $S$.    

\begin{figure}[t]
\centerline{\includegraphics[width=0.9\linewidth]{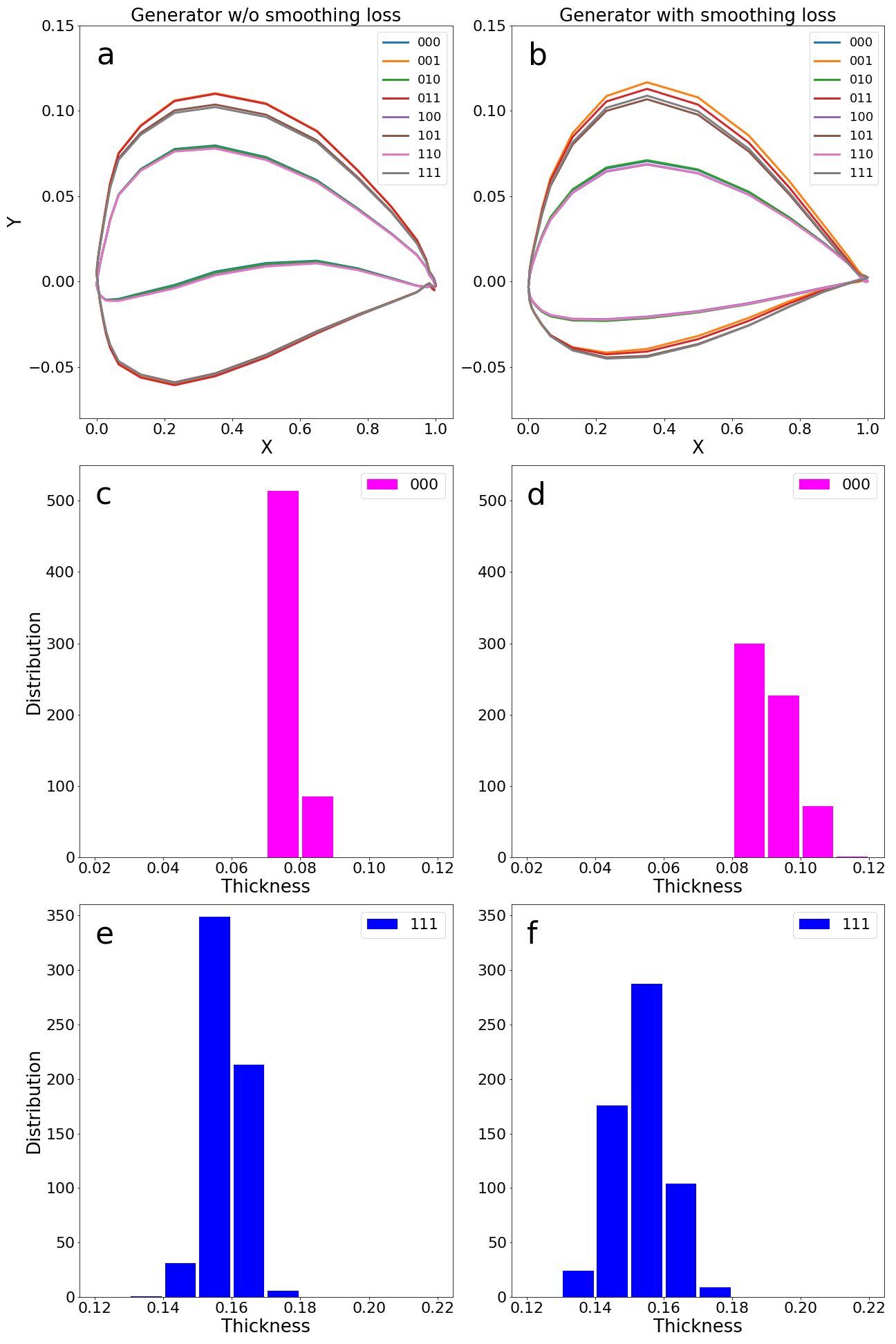}}
\caption{Comparison of airfoils which are generated from GAN trained without smoothing loss and augmented with a post-processing smoothing filter (3 panels in left column), and GAN trained with smoothing loss (3 panels in right column). (a) and (b) show the mean shape of 600 airfoils, a total of 8 shapes corresponding to 8 classes. Distribution of $\tau$ of 600 airfoils for a fixed class: (c) and (d) for class-000, and (e) and (f) for class-111.}
\label{Fig:Results}
\end{figure}
\begin{table}[htbp]
\caption{Performance of GAN}
\begin{center}
\begin{tabular}{|c|c|c|c|c|c|c|}
\hline
\textbf{Class}&\multicolumn{3}{|c|}{\textbf{w/o Smoothing Loss}} &\multicolumn{3}{|c|}{\textbf{with Smoothing Loss}}\\
\cline{2-7} 
{} & $\text{ACC}^{(\tau)}(\%)$ & $\sigma^{(\tau)}$ & $S$ & $\text{ACC}^{(\tau)}(\%)$ & $\sigma^{(\tau)}$ & $S$ \\
\hline
000 & 100 & 0.007  & 0.003 & 100 & 0.057 & 0.017 \\
001 & 100 & 0.024  & 0.009 & 100 & 0.044 & 0.017 \\
010 & 100 & 0.007  & 0.003 & 100 & 0.061 & 0.018 \\
011 & 100 & 0.028  & 0.01  & 100 & 0.036 & 0.02 \\
100 & 100 & 0.005  & 0.003 & 100 & 0.05  & 0.017 \\
101 & 100 & 0.027  & 0.009 & 100 & 0.044 & 0.02 \\
110 & 100 & 0.008  & 0.003 & 100 & 0.057 & 0.018 \\
111 & 100 & 0.03   & 0.009 & 100 & 0.043 & 0.02 \\
\hline
\end{tabular}
\label{Table:performance}
\end{center}
\end{table}
The comparative performance of these two GAN models is summarized in Table~\ref{Table:performance}. Both models achieve a 100$\%$ accuracy ($\text{ACC}^{(\tau)}$), indicating that the generated airfoils consistently exhibit the correct (high/low) thickness corresponding to their respective classes. 
However, smoothGAN exhibits notably elevated values for both $\sigma^{(\tau)}$ and $S$.
As seen in Fig.~\ref{Fig:Results}, distribution of $\tau$ is broader for smoothGAN than the former GAN leading to approximately 1.5 to 10 times higher $\sigma^{(\tau)}$ values.  
Similarly, smoothGAN exhibits approximately 2 to 6 times higher $S$ values than the former GAN. It is noteworthy that the minimum value of S is zero implying that the model generates only one airfoil shape. This situation may occur in the presence of mode collapse, a common drawback associated with GANs~\cite{yang2019diversity}.

\section{Conclusions}
In this study, we developed a methodology to generate smooth airfoils using GAN. We showed that conditional GAN has limitations in generating smooth curves, supporting the previous findings reported in the literature. Typically, this shortcoming has been addressed by augmenting a smoothing filter with GAN as a post-processing technique. 
We demonstrated explicitly that employing such a post-processing technique can attain high accuracy. Nevertheless, this approach may generate a restricted range of airfoil shapes, and the absence of diversity in shape can constrain the GAN's ability to explore novel designs.

Our proposed methodology integrates a custom loss function with GAN, comprising a smoothing loss that penalizes deviations of the generated shape from the reference shape assessed through a moving average. This approach enhances the GAN's capability to consistently generate smooth airfoils, with 2 to 10 times greater diversity compared to the original GAN.

In future applications, we plan to implement our proposed GAN methodology in various scenarios to generate smooth curves and surfaces. One potential application could be in the realm of 3D airfoil design.

\section*{Acknowledgment}
This research is supported by Agency for Science Technology and Research (A*STAR) under the AME Programmatic project: Explainable Physics-based AI for Engineering Modelling \& Design (ePAI) [Award No. A20H5b0142].

\end{document}